\documentclass[journal]{IEEEtran}
\usepackage{amsmath,amsfonts}
\usepackage{algorithmic}
\usepackage{algorithm}
\usepackage{amsmath}
\usepackage{array}
\usepackage[caption=false,font=normalsize,labelfont=sf,textfont=sf]{subfig}
\usepackage{textcomp}
\usepackage{stfloats}
\usepackage{url}
\usepackage{verbatim}
\usepackage{graphicx}
\usepackage{cite}
\usepackage{makecell}
\usepackage{float}
\usepackage{CJK}
\usepackage{multirow}
\usepackage{xcolor}
\usepackage{authblk}
\usepackage{soul}
\usepackage{color}
\begin{document}

\title{Predictive Coding Based Multiscale Network \\ with Encoder-Decoder LSTM for Video Prediction}

\author[1]{Chaofan Ling, Junpei Zhong, \IEEEmembership{Member, IEEE}, Weihua Li, \IEEEmembership{Senior member, IEEE}
\thanks{C. Ling and W. Li are with South China University of Technology, Shine-Ming Wu School of Intelligent Engineering, Guangzhou, China. J. Zhong is with The Hong Kong Polytechnic University, Department of Rehabilitation Sciences, KLN, Hong Kong. J. Zhong and W. Li are 
Corresponding authors: \\ Junpei Zhong: joni.zhong@polyu.edu.hk  \\ Weihua Li: whlee@scut.edu.cn}}

\maketitle

\begin{abstract}
We present a multi-scale predictive coding model for future video frames prediction. Drawing inspiration on the ``Predictive Coding" theories in cognitive science, it is updated by a combination of bottom-up and top-down information flows, which can enhance the interaction between different network levels. However, traditional predictive coding models only predict what is happening hierarchically rather than predicting the future. To address the problem, our model employs a multi-scale approach (Coarse to Fine), where the higher level neurons generate coarser predictions (lower resolution), while the lower level generate finer predictions (higher resolution). In terms of network architecture, we directly incorporate the encoder-decoder network within the LSTM module and share the final encoded high-level semantic information across different network levels. This enables comprehensive interaction between the current input and the historical states of LSTM compared with the traditional Encoder-LSTM-Decoder architecture, thus learning more believable temporal and spatial dependencies. Furthermore, to tackle the instability in adversarial training and mitigate the accumulation of prediction errors in long-term prediction, we propose several improvements to the training strategy. Our approach achieves good performance on datasets such as KTH, Moving MNIST and Caltech Pedestrian.  Code is available at https://github.com/Ling-CF/MSPN
\end{abstract}

\begin{IEEEkeywords}
Video frame prediction, predictive coding, adversarial training.
\end{IEEEkeywords}

\section{Introduction}

\begin{figure}[]

	\centering{\includegraphics[width=3.2in]{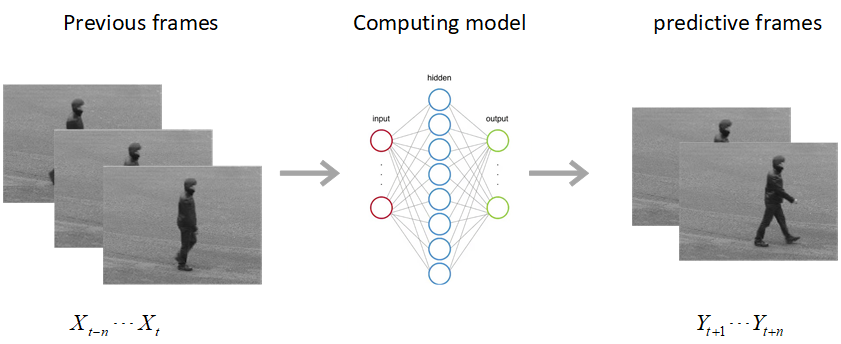}}
	\caption{Computational model learns to predict future frames ($Y_{t+1}, ..., Y_{t+m}$) by analyzing spatial-temporal relationships in the previous video frames ($X_{t-n}, ..., X_{t}$)}
	\label{fig:visual prediction}
\end{figure}

Video prediction is a pixel-intensive task that predicts and generates future frames by learning from historical frames (Fig \ref{fig:visual prediction}). This looking-ahead ability can be applied to various fields such as autonomous driving \cite{morris2008learning, hu2020probabilistic}, robot navigation \cite{finn2017deep}, activities and events prediction \cite{kitani2012activity, shi2017deep, bhattacharyya2018long}. For instance, it endows the self-driving car to forecast changes in the traffic situation to make actions beforehand.  Additionally, as a self-supervised learning task, it can be also transferred to tasks such as video classification \cite{han2019video, wang2020self}, to mitigate the difficulty of label acquisition in supervised learning.

However, predicting how future scenarios will unfold is a challenging task. Existing works still suffer from a lack of appearance details or low prediction accuracy \cite{shi2015convolutional, villegas2017decomposing, wang2018predicting, lin2020motion, li2020video, tan2023temporal, chang2021mau, ye2023video}. Although employing different convolutional and recurrent units, most of them basically share similar hybrid architectures. A commonly used framework is the Encoder-RNNs-Decoder network. It is characterized by first performing encoding on the input frames to extract high-level representations, which are then used to calculate and predict next representations with the recurrent units. Finally, the decoder reconstructs the pixels based on the prediction. For examples, the VPTR \cite{ye2023video}, TAU \cite{tan2023temporal} and MAU \cite{chang2021mau} all adhere to this framework, but the recurrent units used are different. The VPTR utilizes transformer network to perform prediction, whereas TAU and MAU employ their proposed temporal or motion units. Despite its simple calculation and implementation, this framework only performs predictions in high-level semantic spaces, while ignoring short-term modeling of low-level representations. The prediction of future video frames mainly relies on the physical information rather than semantics \cite{oprea2020review}, and only predicting the semantics may cause the predicted frames to repeat the previous frames, or deviate from the physical trajectory.

Another popular framework is stacked ConvLSTMs \cite{shi2015convolutional, shi2017deep, su2020convolutional}, in which they directly integrate the convolution into the recurrent units. It overcomes the limitation of standard LSTM that the spatial correlation is not explicitly modeled. However, its hidden states are propagated from bottom up, and the spatial representations are encoded and decoded level by level. In view of the special architecture of LSTM, the calculation of this framework is actually more complicated. Additionally, its memory state is only updated along the time arrow, exhibiting minimal dependency on the hierarchical visual features of other network levels. To address this problem, PredRNN models \cite{wang2017predrnn, wang2018predrnn++, wang2022predrnn} proposed to propagate the memory state of the top network level to the bottom for the calculation of next time step. Nevertheless, the connection remains inadequate and the computation remains complex.

Drawing inspiration on the ``predictive coding" theories and ``coarse to fine" approach, we propose a multi-scale predictive network (MSPN) to address the above problems. 
In the field of cognitive science, predictive coding is one of the most critical cognitive frameworks \cite{rao1999predictive, friston2010free, friston2006free, han2017rhythms,aitchison2017or ,teufel2020forms}. 
It argues that an internal model in the brain continuously generates and updates a prediction of the environment, which will be used to compare with actual sensory inputs to generate prediction errors. These errors are then used to update and correct the intrinsic representation model. According to Rao et al. \cite{rao1999predictive}, the model is updated through a combination of bottom-up and top-down information flows, which can perform encoding and decoding simultaneously. It connects neurons at different network levels with predictions and prediction errors, where the prediction errors further constitute an efficient feedback connection to better guide the learning of network model. Unfortunately, traditional predictive coding models are only hierarchically predicting what is happening, not the future \cite{hogendoorn2019predictive}. Therefore, we propose to combine the coarse-to-fine approach to improve the architecture. This method of gradually restoring pixels has also been shown to obtain high-quality generated images \cite{karras2017progressive, aigner2018futuregan}. Our contributions can be summarized as follows:

\begin{itemize}
	\item We construct a novel multi-scale predictive coding network for video prediction. It inherits the advantage of  predictive coding models being updated through bottom-up and top-down information flows, but the difference is that it directly predicts the RGB images rather than the activities of lower-level neurons.
	\\~
	\item In light of the challenges observed in the previously mentioned Encoder-LSTM-Decoder framework, we present a solution: the Encoder-Decoder LSTM (EDLSTM) module. This innovative approach seamlessly incorporates the encoder-decoder network into LSTM, thereby empowering the model to effectively merge the memory states and current inputs, to study more reliable temporal and spatial dependencies.
	\\~
	\item In conjunction with an improved training strategy, we achieve remarkable performance on multiple datasets, in which we try to balance the prediction accuracy and visual effects.
	
\end{itemize}

\section{Related Work}

Building on the great success of deep learning, research on video prediction has received more and more attention. Early works focused on directly predicting  pixel intensities by implicitly modeling scene dynamics and underlying information. Inspired by the two-stream architectures for action recognition \cite{simonyan2014two}, video generation from a static image \cite{vondrick2016generating} and unconditional video generation \cite{tulyakov2018mocogan}, some people try to decompose the video into motion and content, process them on separate paths. Subsequently, generative adversarial training has shown encouraging performance in image generation, and adversarial training-based stochastic video prediction methods have also exploded. In addition, methods based on 3D convolution and predictive coding are also worthy of attention. Next, we will briefly classify and summarize the above video prediction methods, and introduce some of the most representative models and methods.

The MCNet \cite{villegas2017decomposing} first proposed the separation of motion and content. The authors encode the local dynamics and spatial layout separately, in which they only predict the motion state and finally fuse with the content to obtain the predicted frame. However, the referred motion is obtained by subtracting the previous frame from the subsequent frame, which only describes the changes at the pixel level. So when the scene is complex and varied, it is disastrous. Lin et al. tried to use channel and spatial attention modules to make the model (MAFENet) focus more on the main part of moving objects \cite{lin2020motion}. The idea is positive, but the essence is still to directly encode the pixel difference between frames. Prior to MAFENet , Wang et al. \cite{wang2018predicting} proposed to only focus on the region of interest (ROI) of a video sequence instead of modeling the entire frame. They extract the ROI by calculating the $l_2$ difference of the entire sequence, which is useful in the case of static background or smooth changes between frames. The disadvantage is that it needs to choose a suitable $l_2$ threshold for different datasets to match different scenarios, which may have a great impact on the results. Li et al. performed separation in the decoding stage, predicting optical flow and appearance separately. Compared with directly predicting video frames, this method achieves higher accuracy, but also increases the computational overhead \cite{li2020video}. The DDPAE \cite{hsieh2018learning} proposed to decompose the high-dimensional video into components
represented with low-dimensional temporal dynamics. The authors decomposed digits on the Moving MNIST dataset into appearance and spatial location, so that only the latter needs to be predicted. However, this approach does not seem to be suitable for complex natural scenes. 

Generative adversarial networks have shown superior performance in image generation. The first video prediction method trained in an adversarial style was proposed by Mathieu et al. \cite{mathieu2016deep}. They designed a novel image gradient-based GDL regularization loss to replace the mean square error MSE to generate sharp video frames. Lee et al. proposed a typical stochastic adversarial video prediction method \cite{lee2018stochastic}. To address the problem of deterministic models averaging the future into a single, ambiguous prediction, the authors proposed using variational network to model the underlying randomness, and combined with adversarial training to generate clear visualizations. The reported predicted image quality is good, but it underperforms on quantitative evaluations such as SSIM and PSNR. Simulating the human intelligence system is another popular way. Inspired by the characteristics of the frequency band decomposition of the human visual system, Jin et al. proposed to incorporate wavelet analysis into the prediction network \cite{jin2020exploring}, and cooperate with adversarial training to sharpen the image. The specific operation is to decompose the input video frame into multiple frequency band inputs through discrete wavelet transform. The idea is similar to content and motion separation, except that they are separated from the perspective of frequency.

3D convolutions can effectively preserve local dynamics and model short-term features. But it suffers from poor flexibility (for example, the dimension are usually fixed, and the number of input or output frames cannot be adjusted casually), difficult training and long-distance video reasoning. Therefore, some people attempted to combine 3D convolution and recurrent neural networks to improve the flexibility of learning and reasoning while preserving short-term features. E3D-LSTM is a typical 3D video prediction model proposed by Wang et al. \cite{wang2018eidetic}. They divided the input video sequence into several groups, and each group contains $T$ video frames. Then, they used 3D convolutions for video group encoding, which are fed into LSTM nested with 3D convolutions for prediction. The 3D convolution is responsible for learning the short-term memory within each group, and the LSTM performs long-term modeling to effectively manage long-term historical memory. The disadvantage is that the training is difficult and the calculation overhead is large—every time a video frame is predicted, a set of video sequences with a length of $T$ must be input and calculated.

The ability to predict and reason about future events is the essence of intelligence. Actually, early works \cite{softky1996unsupervised, hollingworth2004constructing} on future prediction were inspired by the predictive coding paradigm . One of the most influential works on video prediction based on predictive coding is the PredNet model proposed by Lotter et al. \cite{lotter2017deep}. It used vertically stacked convolutional LSTM (ConvLSTM) \cite{shi2015convolutional} as the main carrier, and strictly followed the computing style of the traditional predictive coding framework: lower-level neurons transmit prediction errors upward, and higher-level neurons propagate predictions downward. The presented results are superior to the contemporaneous work \cite{mathieu2016deep}, but they only explored the prediction results of the next frame and did not report the performance on long-term prediction. Similarly, Elsayed et al. \cite{elsayed2019reduced} also used ConvLSTM to construct a predictive coding based video prediction model. In order to reduce the computational overhead, they tried to reduce the model parameters and training time by reducing the gating calculation in LSTM, but they also only reported the prediction results of the next frame. Straka et al. constructed a video prediction model named PrecNet by imitating the PredNet model \cite{straka2020precnet}. They changed the direction of the information flow—propagating prediction upward and prediction error downward, which is the opposite of how predictive coding is computed, but the authors did not report the reason.

In addition to the above methods,  Hu et al. \cite{hu2023dynamic} proposed a dynamic multi-scale network to model the motion at different scales by dynamic optical flow estimation. It can adaptively select sub-networks according to the input frames for efficient prediction. By drawing inspiration from transformer-based video generation \cite{yan2021videogpt}, Xi et al. proposed to use transformer network to perform prediction \cite{ye2023video}. However, their proposed model exhibited a similar architecture to Encoder-LSTM-Decoder, with LSTM being replaced by a transformer. Such models only perform predictions in the high-level semantic space, disregarding most of the low-level details. This is why we propose to embed the encoder-decoder network into the LSTM. According to Oprea et al. \cite{oprea2020review}, successive video frames exhibit variations in pixel space, yet maintain consistency in semantic space. Therefore, the Encoder-LSTM-Decoder network is more suitable for predictive tasks that only need to generate video frames that conform to the current semantics. For example, different from the short-term predictions based on physical information, Li et al. first applied video prediction to instructional task, who focus more on the comprehension of semantics between step order and task \cite{li2022order}. 

\begin{figure}[!b]
	\centering{\includegraphics[width=0.495\textwidth]{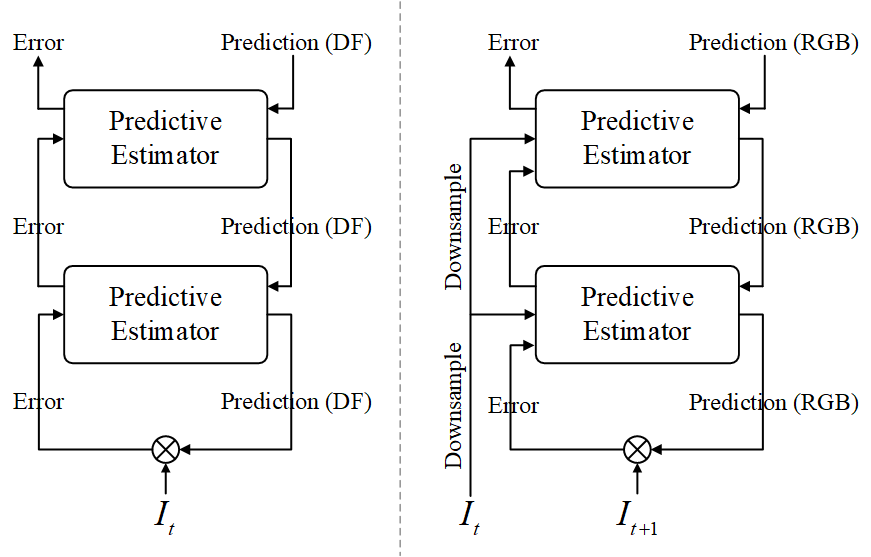}}
	\caption{Differences between traditional predictive coding framework (left) and our model (right). In the figure, $I_t$ denotes the input at time step $t$, ``DF" denotes ``Deep Feature" and ``RGB" denotes a RGB image.}
	\label{fig:PC-MSPN}
\end{figure}

\section{Methods}

\begin{figure*}[!t]
	\centering{\includegraphics[width=1.0\textwidth]{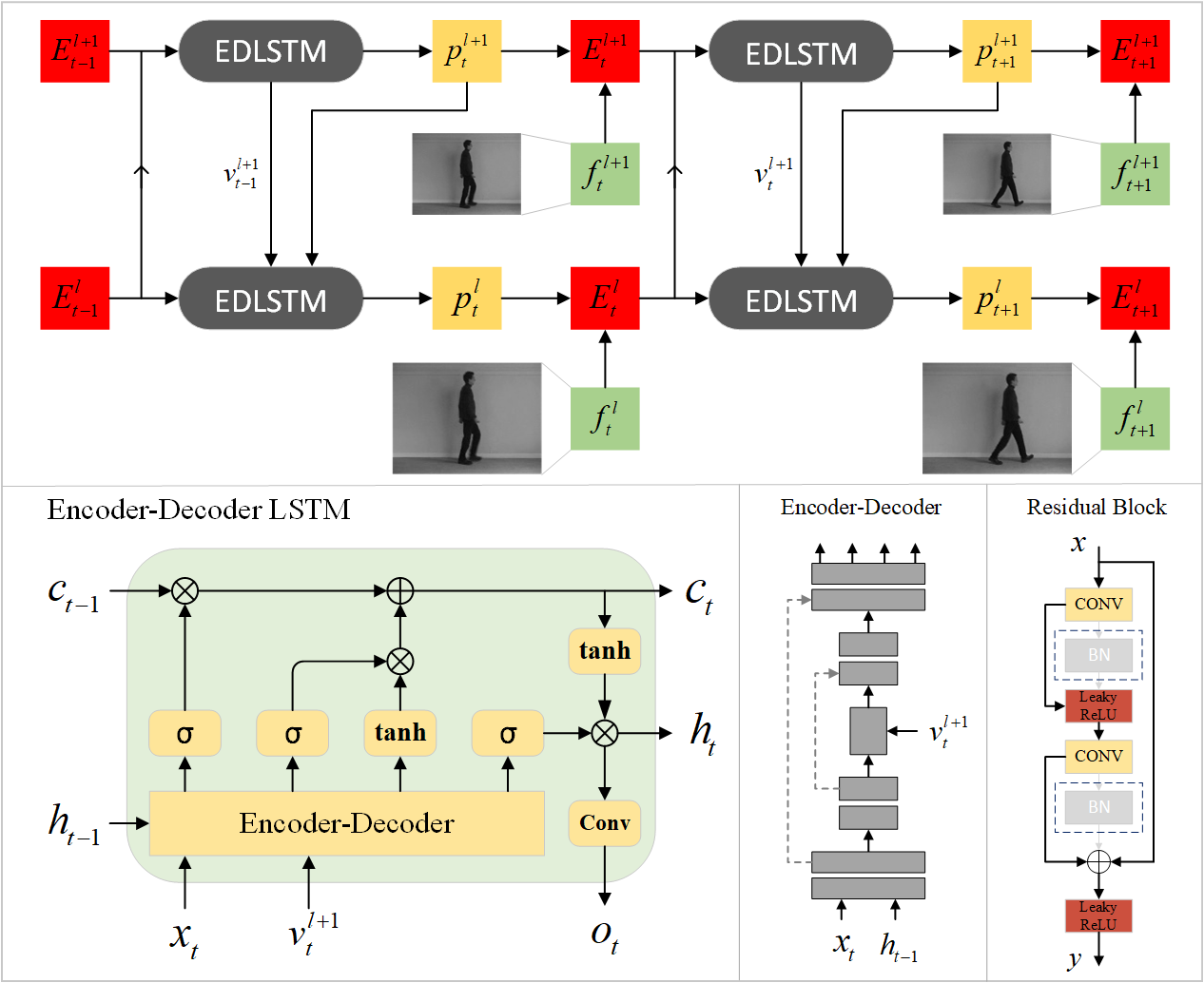}}
	\caption{The overview of multi-scale predictive network architecture, where $P_t^l$, $E_t^l$, $f_t^l$ denote the predicted RGB image, prediction error and the input frame at time-step $t$ and level $l$, respectively. At the bottom, the figure shows the detailed structure of encoder-decoder LSTM module, where $v_t^{l+1}$ represents the encoded high-level semantic information from higher level, $o_t$ denotes the output RGB image. The definition of other parameters is the same as the regular LSTM network.}
	\label{fig:MSPN}
\end{figure*}

\subsection{Network Model}

First, we make improvement to the traditional predictive coding model (PCM) to design the overall framework of our network. As shown in the left of Figure \ref{fig:PC-MSPN}, the predictive coding framework is updated and corrected by the combination of top-down and bottom-up information flows, which performs encoding and decoding simultaneously at each network level. In this hierarchical model, neurons of each level strive to predict the subsequent activity of lower level neurons. These predictions are then transmitted to the lower level and compared with the actual activity of local neurons, resulting in the generation of prediction errors, which will be fed back to higher-level neurons to update their state. 

Drawing inspiration on PCM, our model is also calculated in the predictive coding style. Compared with traditional stacked convolutional LSTM models, it can not only enhance the interaction between different network levels, but also reduce computational overhead. However, the traditional PCM is only hierarchically predicting what is happening, not the future \cite{hogendoorn2019predictive}. To address the problem, we propose to combine the coarse-to-fine approach to improve the architecture to match the task of future video frame prediction. As shown in the right of Figure \ref{fig:PC-MSPN}, neurons at each level in our model directly predict external stimuli (RGB images) rather than the activities of lower-level neurons, in which higher levels neurons predict lower-resolution RGB images, while lower levels predict higher-resolution RGB images.  

Additionally, we propose to combine the stimulus of previous time step ($I_t$) to directly predict subsequent stimulus($I_{t+1}$) instead of activities of lower level neurons. Actually, the external stimuli which contains more useful low-frequency information is also necessary for the predictive unit, to facilitate the convolutional network to study the internal representation and synthesize future frames. Deep neural networks have been shown to be more inclined to study low-frequency signals than sparse prediction errors \cite{xu2019frequency, luo2022upper}.  More implementation details will be described below.

\textbf{Network Architecture} \quad The overall architecture of multi-scale predictive network is shown in Fig \ref{fig:MSPN}, which uses Encoder-Decoder LSTM (EDLSTM) as the main carrier and follows a strict top-down and bottom-up propagation manner of information flow. In the figure, $f_t^l$ denotes the RGB image at time-step $t$ and level $l$, which is obtained by directly downsampling the original input frame. In the actual calculation, we obtain the RGB image of each level through $2 \times$ downsampling, whose size can be defined as $(B, 3, \frac{H}{2^l}, \frac{W}{2^l})$, where $B$, $H$ and $W$ denote batch-size, high and width of the original input, respectively. The $P_t^l$ denotes prediction made by neurons at level $l$ and time-step $t$, its purpose is to predict the RGB image $f_t^l$. By computing the difference between the prediction $P_t^l$ and RGB image $f_t^l$, we can obtain the prediction error $E_t^l$. Similar to PredNet \cite{lotter2017deep}, it is defined as the collective of positive and negative errors $E_t^l = C(ReLU(f_t^l-P_t^l), ReLU(P_t^l-f_t^l))$, where $C$ represents concatenating two tensors together by channel $(B, 6, \frac{H}{2^l}, \frac{W}{2^l})$. This design has its biological basis—there are two populations of error computing neurons, one of which is dedicated to positive errors and the other to negative errors \cite{rao1999predictive}.The prediction error is used to update and correct the internal representation of  neurons. When the scene is predictable, its signal is sparse, so the predictive coding model is efficient.

In general, the EDLSTM units in each level receive three kinds of inputs: predictions from higher level $P_t^{l+1}$ (except the highest level), prediction errors from the local level $E_{t-1}^l$ and lower level $E_{t-1}^{l-1}$ (except the lowest level), to make local prediction $P_t^l$. Since the input signal is at the pixel level, it is necessary to use a codec network to realize the encoding and decoding of information. Different from the traditional Encoder-LSTM-Decoder model, we directly encapsulate an Encoder-Decoder network which is connected in a skip-connection manner into the LSTM network(Fig \ref{fig:MSPN}) to enhance the predictive ability. In the actual calculation, it first concatenates the current input signals $x_t$ ($x_t$ represents the collective of the aforementioned input signals including prediction, sensory input and prediction errors) and previous hidden state $h_{t-1}$, and then uses the codec network for end-to-end calculation. Next, we split the output of the codec network into 4 kinds of gating signals of LSTM network (Eq.\ref{split}, where $f_t, i_t, o_t, \hat{c}_t$ denote forgetting factor, input factor, output factor, and short-term memory), followed by different activation (Eq.\ref{activate}) and LSTM gating calculation (Eq. \ref{LSTM Gat}). Finally, the output RGB image is obtained by performing another convolutional calculation on the current hidden state $h_t$. 

\begin{equation}
f_t, i_t, o_t, \hat{c}_t = Split(Codec(x_t, h_{t-1}))
\label{split}
\end{equation}
\begin{equation}
	(f_t, i_t, o_t), \hat{c}_t = Sigmoid((f_t, i_t, o_t)), Tanh(\hat{c}_t)
	\label{activate}
\end{equation}
\begin{equation}
c_t = f_t \cdot c_{t-1} + i_t \cdot \hat{c}_t, \ \ h_t = o_t \cdot Tanh(c_t)
\label{LSTM Gat}
\end{equation}

\begin{figure}[!t]
	\centering{\includegraphics[width=0.49\textwidth]{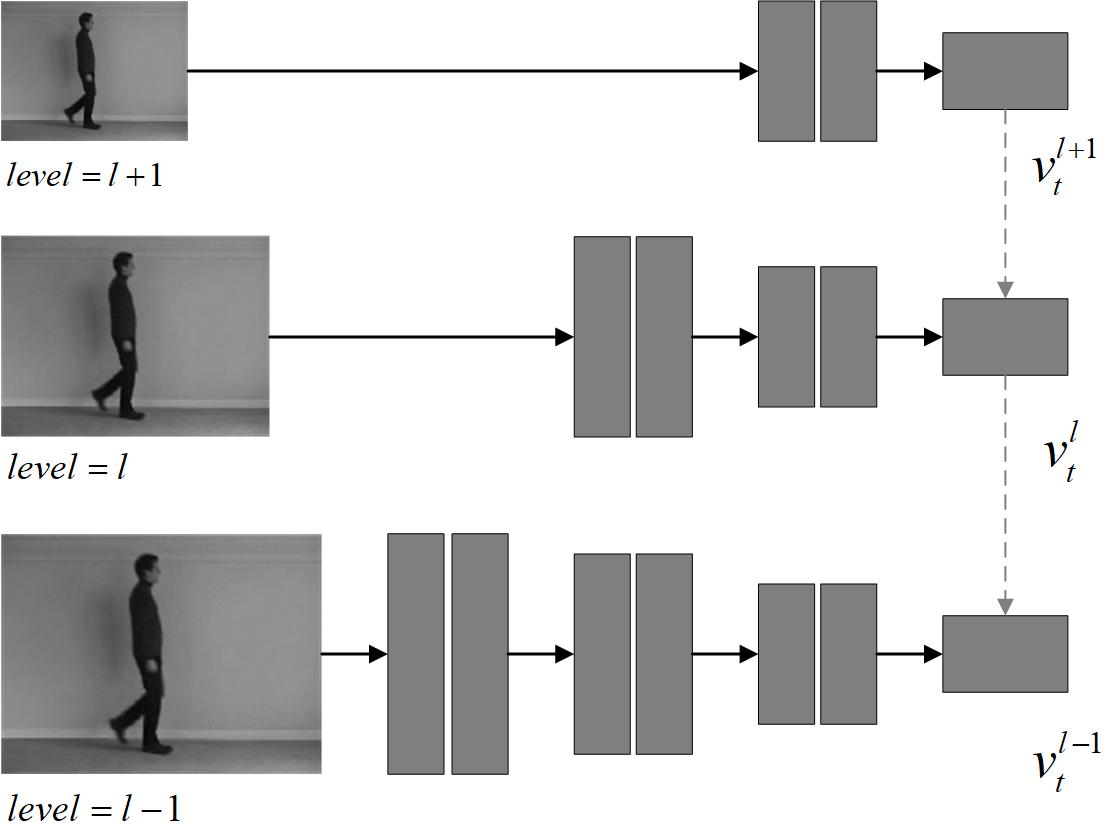}}
	\caption{We share the high-level semantic information encoded by higher-level neurons. The $v_t^l$ denotes the final encoded features at time-step $t$ and level $l$. We reduce the number of network layers in higher-level, to ensure that the input of each level is encoded into features of the same size (different depth (channels), and the depth of higher level is smaller) while reducing computational overhead. Then the higher-level encoded features $v_t^{l+1}$ are passed down, and the lower-level encoded features $v_t^l$ are calculated and updated.}
	\label{fig:feature-fusion}
\end{figure}

In contrast to traditional Encoder-LSTM-Decoder frameworks that solely perform predictions in the semantic space, the EDLSTM module enables comprehensive interactive calculation of the LSTM hidden state and current input signals at different levels (pixel and semantic). This leads to the acquisition of more reliable temporal and spatial dependencies. Moreover, the proposed EDLSTM module requires fewer parameters. 
The LSTM hidden state in the EDLSTM module only needs to retain and update the essential information necessary for reconstructing an RGB image, as the final output at each level is a 3-channel RGB image. Consequently, the depth of the hidden state can be reduced (we have set it to 64 channels in this work). The parameters in the EDLSTM module are primarily allocated to the encoder and decoder networks, while the Encoder-LSTM-Decoder framework, not only employs similar encoder and decoder networks, but also entails additional LSTM gating calculations at the semantic level. Since the input signals have already been encoded into a higher-dimensional space by the encoder, a greater number of convolution parameters are necessitated. Based on our ablation experiments, the Encoder-LSTM-Decoder framework requires approximately twice as many parameters as the EDLSTM module.
\begin{equation}
	v_t^l \leftarrow \mathcal{F}_{ \theta } (v_t^l, v_t^{l+1}) 
	\label{eq:variable}
\end{equation}

In addition, as the level increases, the number of covolutional layers of the encoder-decoder networks is reduced  (Fig \ref{fig:feature-fusion}). This approach serves multiple purposes: it helps prevent excessive parameter usage and computational overhead, while ensuring that the encoders at each level encode the input to the same size for seamless sharing of high-level semantic information. To further enhance the connectivity between different network levels, we propose to share the high-level semantic information encoded by higher-level neurons  (Fig \ref{fig:feature-fusion}). In typical multi-scale models, connections are typically limited to pixel-level signals \cite{nah2017deep, tao2018scale}, resulting in weak connections between different levels. As shown in the figure, to address this issue, we propagate down the encoded hidden variable $v_t^{l+1}$ obtained by the higher-level encoder network, which will be combined with the output of local encoder $v_t^{l}$ to update and calculate a new hidden variable (Eq. \ref{eq:variable}, where $\mathcal{F}_{ \theta }(\cdot)$ represents convolutional calculation and $\theta$ denotes the parameters). It allows lower-level networks to leverage visual features and semantic messages learned by higher-level encoders, thereby enhancing information utilization and avoiding redundant calculations by lower-level neurons.
~\\

\begin{algorithm}[!b]
	\caption{Calculation of multi-scale predictive network}
	\begin{algorithmic}[1]
		\STATE \textbf{Required: } video frames $f_0, f_1, ..., f_n$
		\STATE \textbf{Output: } predictive frames $P_0, P_1, ..., P_n$
		\FOR{$t=0$ to $T$}
		\FOR{$l=L$ to $0$}
		\IF{$l==L$}
		\STATE $P_t^l, v_t^l = EDLSTM(E_{t-1}^l, E_{t-1}^{l-1})$
		
		\ELSIF{$l==0$}
		\STATE $P_t^l, v_t^l = EDLSTM(E_{t-1}^l, P_t^{l+1}, v_t^{l+1})$
		
		\ELSE
		\STATE $P_t^l, v_t^l = EDLSTM(E_{t-1}^l, E_{t-1}^{l-1}, P_t^{l+1}, v_t^{l+1})$
		\ENDIF
		\ENDFOR
		\FOR{$l=0$ to $L$}
		\STATE $f_t^l = Downsample(f_t, size=(\dfrac{h}{2^l}, \dfrac{w}{2^l}))$
		\STATE $E_t^l = [ReLU(f_t^l-p_t^l),ReLU(p_t^l-f_t^l)]$
		\ENDFOR
		\ENDFOR  
	\end{algorithmic}
	\label{algorithm:MSPN}
\end{algorithm}

\textbf{Algorithm} \quad The calculation of the multi-scale predictive model is described in Algorithm \ref{algorithm:MSPN}, which is implemented by calculating the bidirectional information flow—we first initialize the prediction errors at each level to zero. In the top-down information flow computation (line 4-11 in Algorithm \ref{algorithm:MSPN}), higher-level EDLSTM units receive the prediction error of previous time-step $E_{t-1}^l$, the prediction $p_t^{l+1}$ from higher level (except the highest level, line 6 in the Algorithm) and the prediction error $E_{t-1}^{l-1}$ from lower level (except the lowest level, line 8 in the Algorithm) to make local prediction $P_t^l$, and then propagate down the prediction $P_t^l$ and the encoded latent variable $v_t^l$. Next, in the bottom-up computational flow (line 13-16 in Algorithm \ref{algorithm:MSPN}), we calculate the prediction targets $f_t^l$ for higher levels by downsampling the input frame, followed by calculating the difference between targets and predictions and updating the prediction error $E_t^l$.

\subsection{Training strategies}

\textbf{Training Loss} \quad We directly use the Euclidean distance instead of the mean squared error as the loss function. The mean square error is divided by the total number of pixels on the basis of square error, which can easily cause the gradient to be too small or even disappear, and eventually lead to blurring of the generated image. The computation is defined as Eq. \ref{eq:loss}:
\begin{equation}
	\mathcal{L}_{pix} =  \sum_{t=0}^{T} \sum_{l=0}^{L} \lambda_t \lambda_l \parallel Y_t^l-\hat{Y}_t^l \parallel^2
	\label{eq:loss}
\end{equation}
where $Y_t^l$ and $\hat{Y}_t^l$ denote the ground truth frame and predicted RGB image respectively at level $l$ and time-step $t$. $\lambda_t$ and $\lambda_l$ are the weighting factors by time and network level.

In order to solve the ambiguity caused by the deterministic loss function, the adversarial training is also introduced to sharpen the predicted frame. We try to build a discriminator which also uses residual block as the main carrier to guide the predictive model to generate more realistic frames. The adversarial losses are defined as follows:

\begin{equation}
	\mathcal{L}_D = \mathbb{E}_{x \sim p(x)}[D(x)] +\mathbb{E}_{z \sim p(z)}[1-D(G(z))]
\end{equation}

\begin{equation}
	\mathcal{L}_G = \mathbb{E}_{z \sim p(z)}[D(G(z))]
\end{equation}

where G and D denote the generator and discriminator respectively. The discriminator tries to seek out the difference between the generated frame $G(z)$ ($z$ denotes the input signals of generator) and the real frame $x$ to make distinction. When training the discriminator, we maximize the expectations $D(x)$ and $1-D(G(z))$ to make the discriminator distinguish between real frames and  predicted frames as much as possible. On the contrary, we maximize the expectation $D(G(z))$ while training the generator. In particular, we abandon the cross-entropy classification method but directly score the input images, to better guide the learning of generator.
The total loss of generator is designed as the combination of pixel loss and adversarial loss:

\begin{equation}
	\mathcal{L}_{generator} =  \mathcal{L}_{pix} +\lambda \mathcal{L}_G
	\label{eq:loss total}
\end{equation}

where $\lambda$ denotes the weighting factor. Compared with the adversarial loss, pixel loss is easier to guide the learning of the network and limit its optimization direction, so that the generator will not generate realistic but incorrect frames. However, this limitation may still cause the final generated image to be blurry, so we need  a larger value for $\lambda$.
~\\

\textbf{Improved Adversarial Training} \quad   It is already a consensus that adversarial training is difficult. Intuitively, the generation of an image is much more difficult than its discrimination. So there is often a problem that the discriminator is over-trained, which leads to the disappearance of learning gradient.
In this work, we try to alleviate such a problem by first training the discriminator to preliminarily distinguish between real frames and predicted frames, that is, the score of real frames is higher than that of predicted frames. However, when the score of real frame is much higher than that of the predicted frame, the discriminator is fixed and the generator is trained until the score of  predicted image is close to but still smaller than the real frame. At this point, the generator is fixed and the discriminator is trained, and so on.
In this way, the discriminator is always slightly stronger than the generator, which prevents  the discriminator from overtraining  and  the generator from learning the wrong gradients. 

\begin{algorithm}[!b]
	\caption{Improved adversarial training manner}
	\begin{algorithmic}[1]
		\STATE \textbf{Required: } input signal $z_0, z_1, ..., z_n$, real video frames $f_0, f_1, ..., f_n$, weighting factor $\lambda$, tolerance $c_1, c_2$
		\FOR{$epoch = 0$ to $m$}
		\STATE \textbf{/ Train discriminator /}
		\WHILE{$True$}
		\STATE $P_n = G(z_n)$ 
		\STATE $P_s, R_s = D(P_n), D(f_n)$ 
		\STATE $\mathcal{L}_D = (R_s - 1)^2 + (P_s + 1)^2$
		\STATE backward $\mathcal{L}_D$
		\IF{$P_s < R_s - c_1$}
		\STATE \textbf{Break}
		\ENDIF
		\ENDWHILE
		\STATE \textbf{/ Train generator /}
		\WHILE{$True$}
		\STATE $P_n = G(z_n)$
		\STATE $P_s, R_s = D(P_n), D(f_n)$ 
		\STATE $\mathcal{L}_G = (P_s - 1)^2$
		\STATE $\mathcal{L}_{total} = \mathcal{L}_{pix} + \lambda \mathcal{L}_G$
		\STATE backward $\mathcal{L}_{total}$
		\IF{$P_s > R_s - c_2$}
		\STATE \textbf{Break}
		\ENDIF
		\ENDWHILE
		\ENDFOR
		
	\end{algorithmic}
	\label{algorithm:train}
\end{algorithm}

The total implementation of training is shown in Algorithm \ref{algorithm:train}, where $P_s$, $R_s$ denote scores of predictive frames and real frames respectively. Tolerance $c_1$ and $c_2$ are two coefficients where $c_1$ is less than $c_2$, they are calculated from the scores of real frames $R_s$ (Table \ref{parameters}). During training, the scores of the predicted frames are always kept within a fixed range ($R_s - c_2 < P_s < R_s - c_1$), so that the discriminator is always slightly stronger than the generator, thus better guiding the learning of the generator.

~\\
\textbf{Long-term Prediction} \quad The accumulation of prediction errors in long-term prediction remains a problem that has not been solved fundamentally. The model is supported by the ground truth frames during training, while it is necessary to use the predicted frames as new input for continuous prediction during testing. There is a distance between the predicted frame and ground truth frame, thus resulting in mismatch between training and testing. Moreover, in the calculation of recurrent neural network based on the Markov chain, when a bad result is produced at a certain time step, it will inevitably affect the subsequent learning. Therefore, with the accumulation of prediction errors, the predicted frames will become increasingly blurry, which is more pronounced when using deterministic prediction methods, mainly manifested as motion blur, ghosting, etc. To mitigate this issue, we propose to adjust the training strategies.

\begin{figure}[!t]
	\centering{\includegraphics[width=0.48\textwidth]{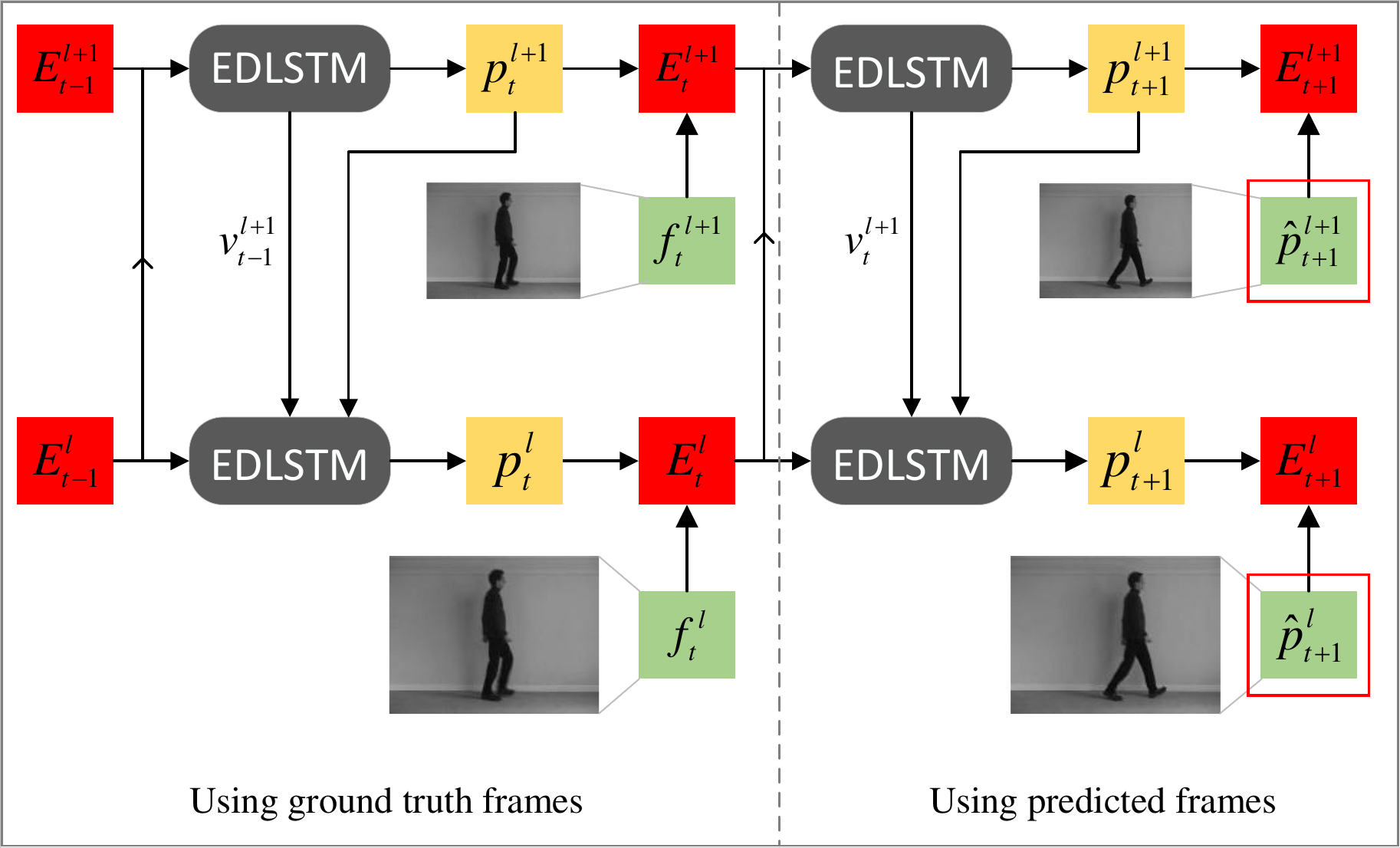}}
	\caption{The model is trained to learn to make prediction continuously by serving predicted frames as new inputs. $\hat{P}_{t+1}^{l}$ denotes the RGB input image of  level $l$ downsampled from the original predicted frame $P_{t+1}^0$.}
	\label{fig:MSPN22}
\end{figure}

As depicted in Fig \ref{fig:MSPN22},  we also use predicted frames as input  during training, trying to train the model effectively extract valuable information from predicted frames while mitigating the impact of prediction errors. We first use $T$ consecutive ground truth video frames as input for calculation to obtain the next predicted frame, which is then served as new input frame for the subsequent calculation. Note that higher-level input frames $\hat{P}_{T+1}^l$ are obtained by directly downsampling the predicted frame $P_{T+1}$ instead of local prediction $P_{T+1}^l$. Specifically, at time step $T+1$, the higher-level EDLSTM units utilize information of previous step to generate predictions, which is then propagated down to gradually restore the original resolution, that is, the final output predicted frame $P_{T+1}$. Next, it is down-sampled (interval sampling or pooling) and propagated to higher levels to calculate and update the prediction errors for the subsequent iteration of computation. Note that the calculation of  loss is still based on the real frames.

\section{Experiments}
In this section, we will first introduce the details of the experimental setup, including the quantitative evaluation metrics used in the task of video prediction and the settings of hyperparameters. Next, we will compare with the state-of-the-art methods on the five commonly used datasets of Moving MNIST \cite{srivastava2015unsupervised}, KTH \cite{schuldt2004recognizing}, Human3.6M \cite{ionescu2013human3}, Caltech \cite{Dollar2012PAMI} and KITTI \cite{Geiger2013IJRR}. We try our best to preprocess the datasets as closely as existing methods  \cite{straka2020precnet, jin2018varnet, mathieu2016deep, lotter2017deep}, and directly make comparison with the demonstrated results. Finally, we conduct an ablation study on the training strategy proposed in this paper and whether to use batch normalization to highlight the effectiveness of our proposed method.

\subsection{Experimental Setup}


\textbf{Quantitative Metrics} \quad Structural Similarity Index Measure (SSIM) and Peak Signal to Noise Ratio (PSNR) are commonly used to evaluate the similarity between two images. The SSIM is calculated from three aspects of luminance, contrast and structure, which is defined as Eq. \ref{eq:ssim}
\begin{equation}
SSIM(x,y)=\dfrac{(2\mu_x\mu_y + c_1)(2\sigma_{xy}+c_2)}{(\mu_x^2+\mu_y^2+c_1)(\sigma_x^2+\sigma_y^2+c_2)}
\label{eq:ssim}
\end{equation}
where $\mu_x, \mu_y$ denote mean of $x, y$ respectively, and $\sigma$ indicate standard deviation. $c_1$ and $c_2$ are two constants used to maintain stability to avoid division by zero. The PSNR is a metric for measuring the image quality between the maximum signal and background noise, which is defined as Eq. \ref{eq:psnr}
\begin{equation}
PSNR = 10 \cdot \log_{10}(\dfrac{MAX_I^2}{MSE})
\label{eq:psnr}
\end{equation}
where $MSE$ denotes the \textit{mean square error} and $MAX_I^2$  indicates the possible maximum pixel value of the images. Besides, we also introduce a recent popular indicator called Learned Perceptual Image Patch Similarity (LPIPS) \cite{zhang2018unreasonable} to measure the difference between two images. It focuses on using deep features to measure perceptual similarity, which is more in line with human perception compared to the traditional metrics. Higher scores of SSIM and PSNR, and lower scores of LPIPS indicate better results.
~\\

\textbf{Experimental Setting} \quad We use PyTorch as platform and Adam \cite{kingma2014adam} as optimizer to implement the algorithm described above. The length of input sequence is fixed at 10 for training and testing, and the number of network levels is set to 4. Other hyper-parameters and introductions are shown in Table \ref{parameters}.

\begin{table}[!t]
	\renewcommand\arraystretch{1.6}
	\small
	\caption{Setting of hyperparameters for training, where $\lambda_t$, $\lambda_l$ and $\lambda$ denote the weighting factors described in Eq. \ref{eq:loss} and Eq. \ref{eq:loss total}. ``$L$" indicates the number of network levels.}
	\begin{tabular}{ccc}
		\hline
		\textbf{parameters} & \textbf{values}           & \textbf{introductions}                        \\ \hline
		$lr_g$        & $1\times 10^{-3}$            & learning rate of generator           \\
		$lr_d$        & $10^{-7} \sim 10^{-9}$          & learning rate of discriminator       \\

		$\lambda_t$    & \makecell[c]{$0, t=0$ \\ $1, t>0$} & weighting factor by time             \\
		$\lambda_l$    & $(L-l)/L$               & weighting factor by level     \\
	    $\lambda$ & 100 & weighting factor of generator loss \\
	    $c_1$ &  $|R_s| / 100$ & upper tolerance \\
	    $c_2$ &  $|R_s| / 50$ & lower tolerance
		\\ \hline      
	\end{tabular}
\label{parameters}
\end{table}

\subsection{Experimental results}

\begin{table}[!t]
	\centering
	\renewcommand\arraystretch{1.41}
	\small
	\caption{Quantitative evaluation on the KTH dataset. The metrics are averaged over the predicted frames. Red and Blue indicate the best and second best results, respectively. }
	\label{Tab:KTH}
	\resizebox{\linewidth}{!}
	{
		\begin{tabular}{lcccccc}
			\hline
			\multirow{2}{*}{Methods} & \multicolumn{3}{c}{10 $\rightarrow$ 20} & \multicolumn{3}{c}{10 $\rightarrow$ 40} \\
			& SSIM $\uparrow$  & PSNR $\uparrow$  & LPIPS $\downarrow$ 
			& SSIM $\uparrow$  & PSNR $\uparrow$  & LPIPS $\downarrow$  \\
			\hline
			MCNet  \cite{villegas2017decomposing}                  & 0.804  & 25.95  & -     & 0.73   & 23.89  & -      \\
			fRNN   \cite{oliu2018folded}                  & 0.771  & 26.12  & -     & 0.678  & 23.77  & -      \\
			PredRNN  \cite{wang2017predrnn}                & 0.839  & 27.55  & -     & 0.703  & 24.16  & -      \\
			PredRNN++   \cite{wang2018predrnn++}             & 0.865  & 28.47  & -     & 0.741  & 25.21  & -      \\
			VarNet  \cite{jin2018varnet}                 & 0.843  & 28.48  & -     & 0.739  & 25.37  & -      \\
			SAVP-VAE \cite{lee2018stochastic}  & 0.852  & 27.77 & \textcolor{blue}{8.36} &0.811  & 26.18  & \textcolor{red}{11.33} \\
			E3D-LSTM \cite{wang2018eidetic}                & 0.879  & 29.31  & -     & 0.810  & 27.24  & -      \\
			STMF  \cite{jin2020exploring}                 & 0.893  & 29.85  & 11.81 & 0.851  & 27.56   & 14.13  \\
			Conv-TT-LSTM  \cite{su2020convolutional}             & \textcolor{blue}{0.907}  & 28.36  & 13.34 & 0.882  & 26.11  & 19.12 \\
			LMC-Memory  \cite{lee2021video}                 & 0.894  & 28.61 & 13.33 & 0.879  & 27.50   & 15.98  \\
			PPNet  \cite{ling2022pyramidal}                 & 0.886  & 31.02  & 13.12 & 0.821  & 28.37  & 23.19  \\
			VPTR  \cite{ye2023video}                 & 0.879  & 26.96  & 8.61 & -  & -  & -  \\
			SimVP \cite{gao2022simvp}      &  0.905      &  \textcolor{blue}{33.72 }     & -    &  \textcolor{blue}{0.886}      &    \textcolor{blue}{32.93}    &   -    \\ 
			TAU \cite{tan2023temporal}      &  \textcolor{red}{0.911}      &  \textcolor{red}{34.13 }     & -    &  \textcolor{red}{0.897}      &    \textcolor{red}{33.01}    &   -    \\
			\hline
			MSPN (ours)      &  0.881      &  31.87     & \textcolor{red}{7.98}     &  0.831      &    28.86    &   \textcolor{blue}{14.04}    \\
			\hline
		\end{tabular}
	}
	
\end{table}

\begin{figure}[!t]
	\centering{\includegraphics[width=0.49\textwidth]{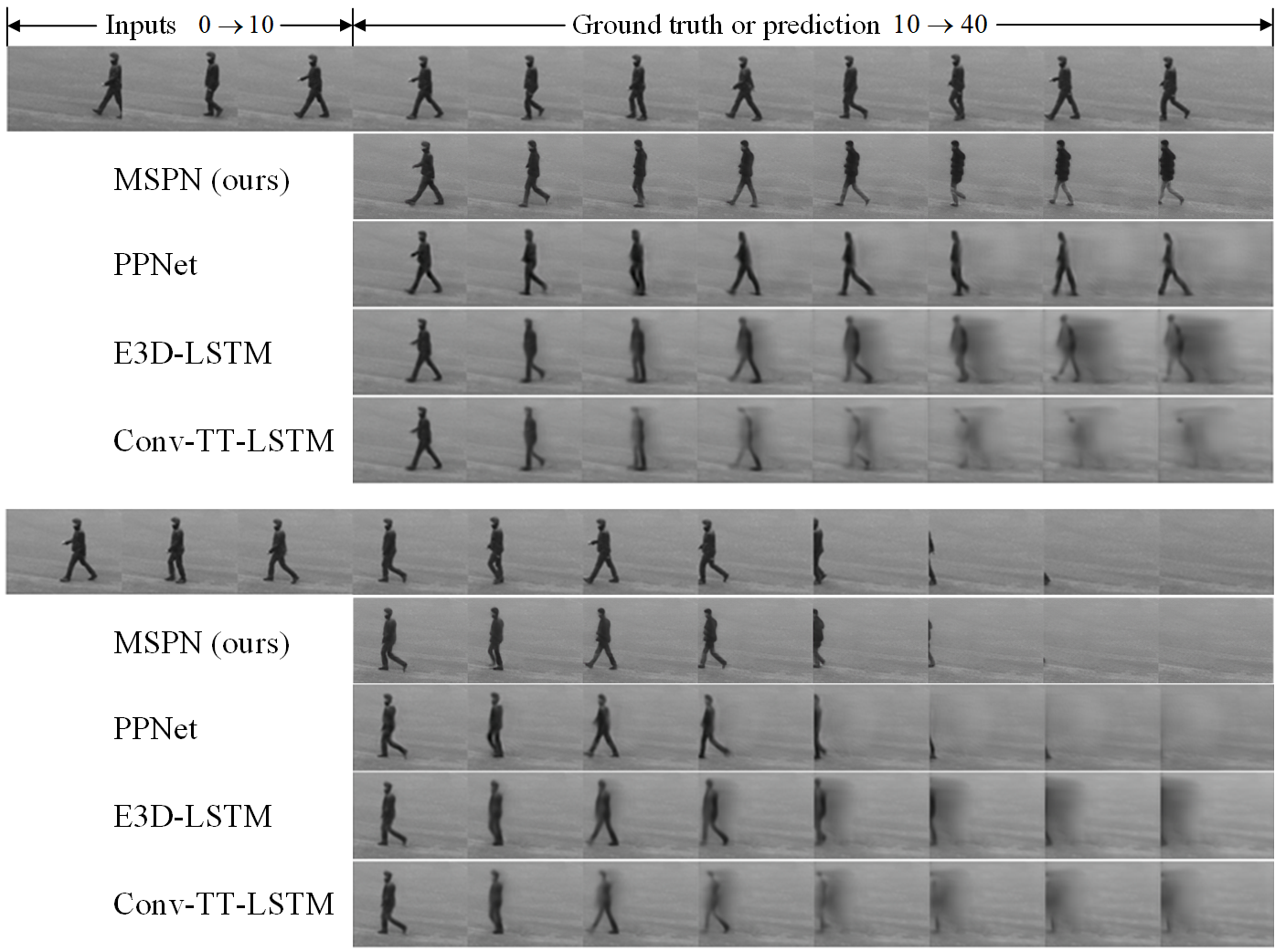}}
	\caption{Visualization examples on the KTH datasets. We use 10 frames as input to predict next 30 frames. The other results are obtained from \cite{ling2022pyramidal}. Zoom in for a better view.}
	\label{fig:KTH}
\end{figure}

\textbf{KTH} \quad  The KTH is widely utilized for video prediction tasks due to its moderate scenes and events. Following the convention of previous works, we allocate person 1-16 for training and person 17-25 for testing. We employ a sequence of 10 frames as input to predict the subsequent 30 frames. The quantitative evaluation results are shown in Table \ref{Tab:KTH}, demonstrating that our method achieves comparable or better performance compared with the state-of-the-art methods. Figure \ref{fig:KTH} shows the visualization examples, highlighting the favorable visual results obtained by our method, while the Conv-TT-LSTM \cite{su2020convolutional}, which obtains a higher SSIM score, exhibits poor performance on the qualitative evaluation. Additionally, the SimVP \cite{gao2022simvp} and TAU \cite{tan2023temporal} achieve even more surprisingly high SSIM and PSNR score, but they do not report the evaluation on perceptual metric LPIPS nor show the visualization examples. The mismatch between quantitative and qualitative assessments remains an unsolved problem in video prediction tasks. A considerable amount of works obtains high pixel-level accuracy but specious results by blurring the future. They prioritize reducing overall prediction errors for better SSIM and PSNR scores, often resulting in blurry images. This is why we propose to combine with adversarial training to sharpen the generated images.
~\\

\begin{table}[]
	\centering
	\renewcommand\arraystretch{1.175}
	\setlength{\tabcolsep}{9pt}
	\tiny
	\caption{Quantitative evaluation on the MNIST dataset. The results are obtained without using adversarial training. The metrics are averaged over the 10 predicted frames. }
	\label{Tab:MNIST}
	\resizebox{\linewidth}{!}
	{
	\begin{tabular}{lcc}
		\hline
		\multirow{2}{*}{Methods} & \multicolumn{2}{c}{10 $\rightarrow$ 20}                               \\
		& \multicolumn{1}{c}{SSIM $\uparrow$} & \multicolumn{1}{c}{MSE $\downarrow$} \\ 
		\hline
		PredRNN++ \cite{wang2018predrnn++}                 & 0.870                    & 47.9                    \\
		E3D-LSTM \cite{wang2018eidetic}                 & 0.910                    & 41.3                    \\

		DDPAE \cite{hsieh2018learning} & -                  & 38.9                   \\
		Conv-TT-LSTM  \cite{su2020convolutional}            & 0.915                    & 53.0                    \\
		LMC-Memory  \cite{lee2021video}               & 0.924                    & 41.5                    \\
		MAU \cite{chang2021mau}            & 0.937        & 27.6          \\
		SimVP \cite{gao2022simvp}            & 0.948        & 23.8          \\
		CrevNet \cite{yu2020efficient}            & 0.949        & 22.3         \\
		TAU \cite{tan2023temporal}            & 0.957        & 19.8 \\
		PredRNN-V2 \cite{wang2022predrnn-v2}            & \textcolor{blue}{0.958}        & \textcolor{blue}{13.4} \\
		\hline
		MSPN (Ours)              & \textcolor{red}{0.963}                    & \textcolor{red}{12.9}   \\ \hline                
	\end{tabular}
}
\end{table}

\begin{figure}[]
	\centering{\includegraphics[width=0.47\textwidth]{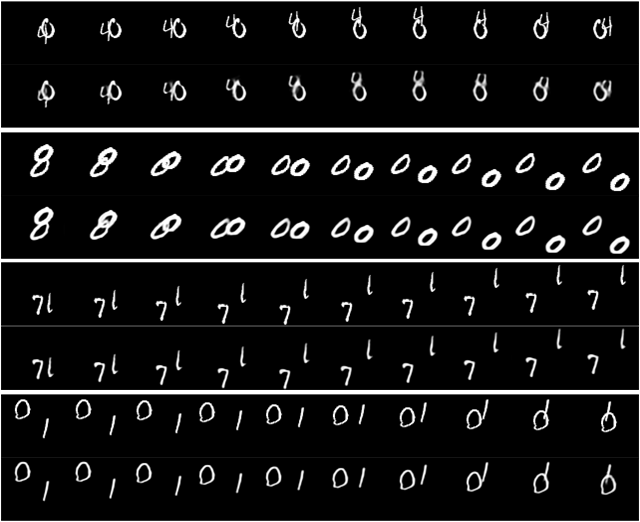}}
	\caption{Visualization examples on the MNIST datasets. The results are obtained without using adversarial training. We use 10 frames as input to predict next 10 frames. In each group, the first row indicates the ground truth frames and the second row indicates predicted frames.  Zoom in for a better view.}
	\label{fig:MNIST}
	\vskip -0.1in
\end{figure}

\textbf{Moving MNIST} \quad The Moving MNIST is an early and widely used synthetic dataset for video representation learning. It consists of simple scenarios and events, where each sequence contains 20 frames depicting the movement of two digits within a $64 \times 64$ box. Similar to previous works, we use 10 frames as input to predict the subsequent 10 frames. Table \ref{Tab:MNIST} shows the quantitative evaluation results on SSIM and MSE. Figure \ref{fig:MNIST} exhibits the visualization examples. Given the simplicity of the scenes and events, we did not employ adversarial training to enhance the sharpness of the generated images to save training overhead. Nevertheless, we can still achieve outstanding performance from the perspective of qualitative evaluation. 
~\\

\begin{table}[]
		\centering
	\renewcommand\arraystretch{1.3}
	\small
	\caption{Quantitative evaluation on the Human3.6M dataset. The best results are marked in bold, where the results of other works are excerpted from \cite{lin2020motion}}
	\label{Tab:Human3.6M}
	\resizebox{\linewidth}{!}
	{
	\begin{tabular}{lcccccc}
		\hline
		Methods                      & Metric & T=2    & T=4    & T=6    & T=8    & T=10   \\ \hline
		\multirow{3}{*}{MCNet \cite{villegas2017decomposing} }       & PSNR   & 30.0   & 26.55  & 24.94  & 23.90  & 22.83  \\ 
		& SSIM   & 0.9569 & 0.9355 & 0.9197 & 0.9030 & 0.8731 \\
		& LPIPS  & 0.0177 & 0.0284 & 0.0367 & 0.0462 & 0.0717 \\
		\hline
		\multirow{3}{*}{fRNN \cite{oliu2018folded}}        & PSNR   & 27.58  & 26.10  & 25.06  & 24.26  & 23.66  \\
		& SSIM   & 0.9000 & 0.8885 & 0.8799 & 0.8729 & 0.8675 \\
		& LPIPS  & 0.0515 & 0.0530 & 0.0540 & 0.0539 & 0.0542 \\
		\hline
		\multirow{3}{*}{MAFENet \cite{lin2020motion}}     & PSNR   & 31.36  & 28.38  & 26.61  & 25.47  & 24.61  \\
		& SSIM   & 0.9663 & 0.9528 & 0.9414 & 0.9326 & 0.9235 \\
		& LPIPS  & 0.0151 & \textbf{0.0219} & \textbf{0.0287} & \textbf{0.0339} & \textbf{0.0419} \\
		\hline
		\multirow{3}{*}{MSPN(Ours)} & PSNR   & \textbf{31.95}  & \textbf{29.19}  & \textbf{27.46}  & \textbf{26.44}  & \textbf{25.52}   \\
		& SSIM & \textbf{0.9687} & \textbf{0.9577} & \textbf{0.9478} & \textbf{0.9382} & \textbf{0.9293} \\
		& LPIPS  & \textbf{0.0146} & 0.0271 & 0.0384 & 0.0480 & 0.0571 \\
		\hline
	\end{tabular}
}
\end{table}

\begin{figure}[]
	\centering{\includegraphics[width=0.49\textwidth]{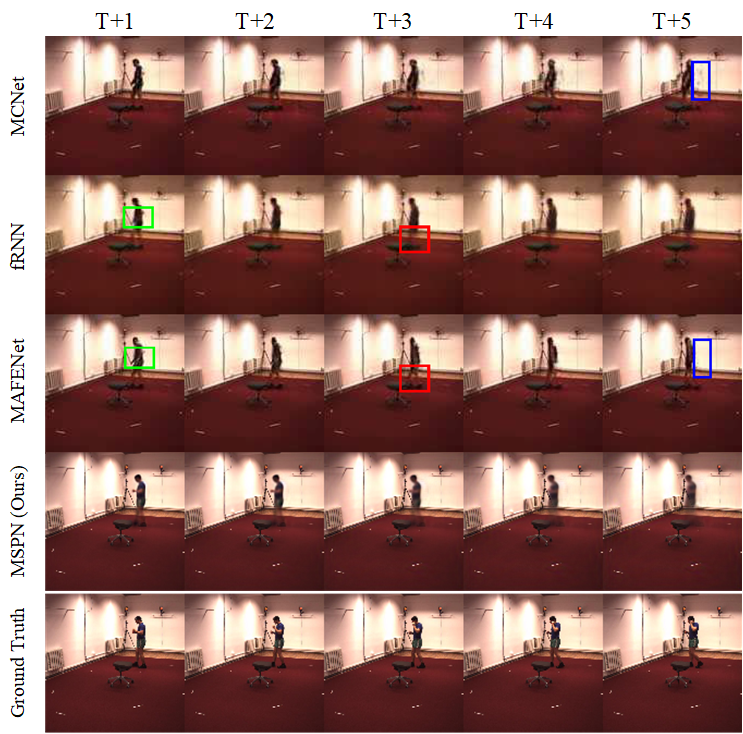}}
	\caption{Visualization examples on the Human3.6M dataset. We use 10 frames as input to predict the next frame. The other results are obtained from \cite{yu2020efficient}. Zoom in for a better view.}
	\label{fig:Human3.6M}
\end{figure}

\textbf{Human3.6M} \quad Human3.6M is a widely used public dataset for 3D human pose estimation research, and it has also been employed in video prediction tasks by various studies due to its time series nature. The dataset comprises videos of performers executing actions within a designated room. Following the approach in \cite{lin2020motion}, we use S11 for testing and the remaining subjects for training. The images are cropped and then adjusted to $128 \times 128$ size. Table \ref{Tab:Human3.6M} and Figure \ref{fig:Human3.6M} show the quantitative evaluation and qualitative evaluation results on this dataset, respectively. Among them, the results of other works are excerpted from \cite{lin2020motion}. Although an exact visualization example matching the one in \cite{lin2020motion} could not be located, we obtained original data with a high resemblance to that example. Thus, figure \ref{fig:Human3.6M} showcases the performance on this set of data. Despite the high accuracy reported in Table \ref{Tab:Human3.6M}, the models seem to struggle in accurately reconstructing the human actions. This is attributed to the dataset's characteristics, where the static background occupies a significant portion of the image sequence. Consequently, the models tend to prioritize learning and reconstructing the static background to achieve high accuracy, often disregarding the human actions depicted in the images.
~\\

\begin{figure}[]
	\centering{\includegraphics[width=0.49\textwidth]{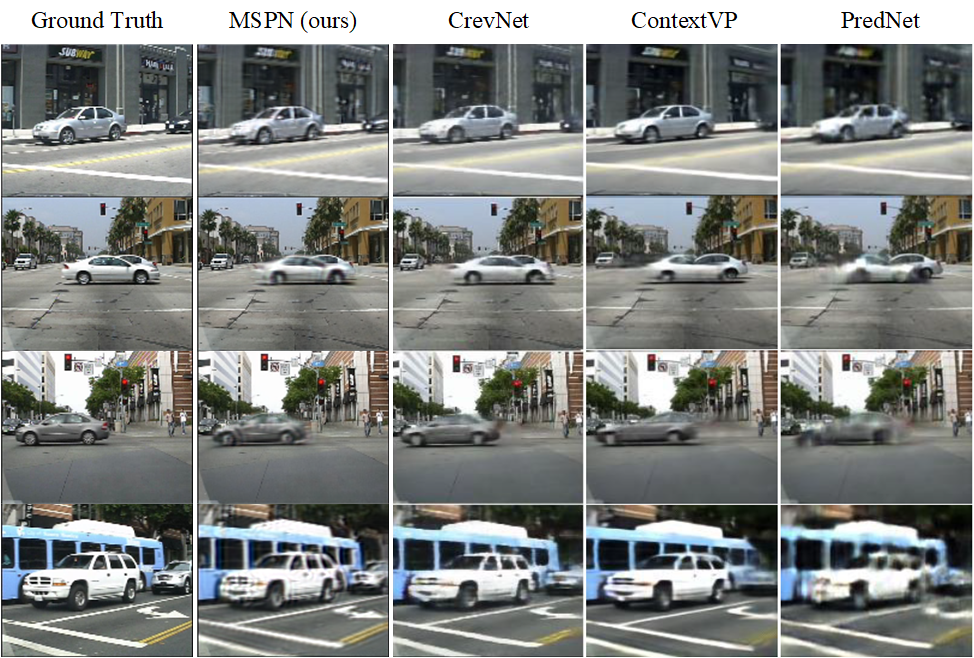}}
	\caption{Visualization examples on the Caltech dataset. We use 10 frames as input to predict the next frame. The other results are obtained from \cite{yu2020efficient}. Zoom in for a better view.}
	\label{fig:Caltech}
\end{figure}

\textbf{Caltech and KITTI} \quad Caltech Pedestrian and KITTI are two other popular datasets, which are obtained by collecting the traffic conditions. In contrast to the above datasets, their scenarios and events are more complex. Among the two, the KITTI dataset exhibits greater variations between frames, making it more challenging for prediction. For comparison and to reduce computational overhead, we resize the images to a resolution of $128 \times 160$. The quantitative evaluation results for the next 5 frames and 10 frames on these datasets are presented in Table \ref{Tab:caltech-kitti}. It is evident that the performance on the KITTI dataset is notably poorer. Nevertheless, we still obtain better results than existing works. Figure \ref{fig:Caltech} shows the visualization examples on the Caltech dataset, demonstrating that our approach can recover more detailed information.  Unfortunately, due to limited results available on the KITTI dataset, we were unable to locate similar visualization examples from the original dataset. As a result, the qualitative evaluation in comparison with other works is not presented in this work.

\begin{table*}[!t]
	\centering
	\renewcommand\arraystretch{1.2}
	\small
	\caption{Quantitative evaluation on the Caltech and KITTI dataset respectively. The metrics are averaged over the 0-5 and 5-10 predicted frames. The results of related works are extracted and computed from \cite{wu2020future}.}
	\label{Tab:caltech-kitti}
	\resizebox{\linewidth}{!}
	{
	\begin{tabular}{lcccccccccccc}
		\hline
		\multirow{3}{*}{Methods}       & \multicolumn{6}{c}{Caltech}                                                                         & \multicolumn{6}{c}{KITTI}                                                                           \\
		& \multicolumn{3}{c}{0-5}                          & \multicolumn{3}{c}{5-10}                         & \multicolumn{3}{c}{0-5}                          & \multicolumn{3}{c}{5-10}                         \\
		& SSIM           & PSNR           & LPIPS          & SSIM           & PSNR           & LPIPS          & SSIM           & PSNR           & LPIPS          & SSIM           & PSNR           & LPIPS          \\  \hline
		\multicolumn{1}{l}{PredNet \cite{lotter2017deep}}    & 0.752          & -              & 36.03          & 0.574          & -              & 68.39          & 0.475          & -              & 62.95          & -              & -              & -              \\
		\multicolumn{1}{l}{MCNET \cite{villegas2017decomposing}}     & 0.705          & -              & 37.34          & 0.488          & -              & 52.92          & 0.555          & -              & 37.39          & -              & -              & -              \\
		\multicolumn{1}{l}{Voxel Flow \cite{liu2017video}} & 0.711          & -              & 28.79          & 0.557          & -              & 44.31          & 0.426          & -              & 41.59          & -              & -              & -              \\
		\multicolumn{1}{l}{Vid2vid \cite{wang2018video}}    & 0.751          & -              & 20.14          & 0.587          & -              & 33.96          & -              & -              & -              & -              & -              & -              \\
		\multicolumn{1}{l}{FVSOMP \cite{wu2020future}}     & 0.756          & -              & 16.50          & 0.592          & -              & 30.06          & 0.608          & -              & \textbf{30.49}          & -              & -              & -              \\
		\hline
		MSPN (Ours)                           & \textbf{0.818}         & 23.88          & \textbf{10.98} & \textbf{0.682} & 19.87 & \textbf{19.56}          & \textbf{0.629}          & 19.44          & 32.10          & 0.447          & 15.69          & 46.52          \\
 \hline
	\end{tabular}
}
\end{table*}

\subsection{Ablation study}

\begin{table}[!t]
	\centering
	\renewcommand\arraystretch{1.175}
	\tiny
	\caption{Ablation studies conducted on network modules using the MNIST dataset. The metrics are averaged over the 10 predicted frames. }
	\label{Tab:MNIST_Ablation}
	\resizebox{\linewidth}{!}
	{
		\begin{tabular}{lccc}
			\hline
			\multirow{2}{*}{Methods} & \multicolumn{2}{c}{10 $\rightarrow$ 20}               &                \\
			& SSIM $\uparrow$ & MSE $\downarrow$ & Params. (M) \\ 
			\hline
			Default                & 0.963                   & 12.9 &      20.5              \\
			w/o EncInfo               & 0.955                   & 16.1          &   16.1        \\
			En-LSTM-De & 0.927                 & 19.2            &   51.5    \\
		   \hline                
		\end{tabular}
	}
\end{table}

\begin{figure}[!t]
	\centering{\includegraphics[width=0.49\textwidth]{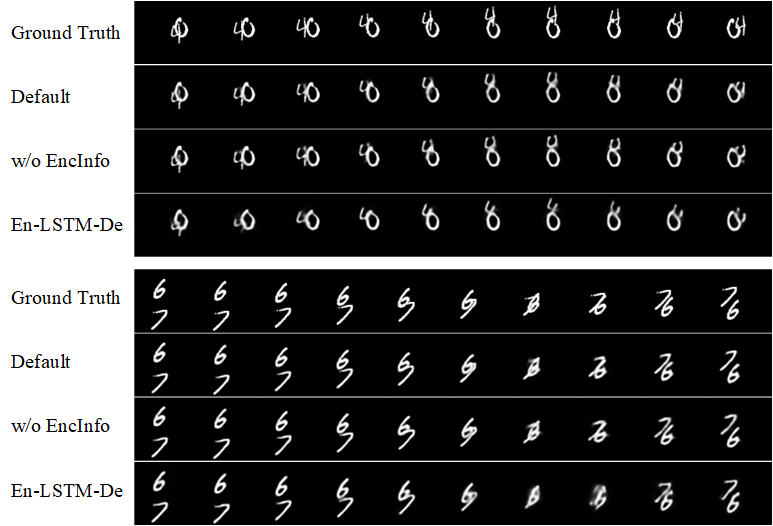}}
	\caption{Visualization examples of ablation studies conducted on network modules using the MNIST dataset. All results are obtained without using adversarial training. Zoom in for a better view.}
	\label{fig:MNIST_Ablation}
\end{figure}

\textbf{Network modules} \quad We conducted a series of ablation studies to assess the impact of the proposed network modules. The results obtained from training on the Moving MNIST dataset are presented in Table \ref{Tab:MNIST_Ablation} and Figure \ref{fig:MNIST_Ablation}. Each ablation was performed by modifying a specific module of the default network.  Firstly, we highlight the superiority of EDLSTM module by comparing with traditional Encoder-LSTM-Decoder (represented as ``En-LSTM-De" in the table and figure) framework. It can be observed that the proposed module outperforms the traditional framework both in quantitative and qualitative evaluation, which even utilize fewer parameters. Secondly, we investigate the significance the of approach for sharing the encoded information from higher-level (``w/o EncInfo" in the table and figure denotes ``without the encoded information from higher-level). Since there is no need to update and calculate a new local encoded signal (that is, Eq \ref{eq:variable}), ``w/o EncInfo" has a smaller number of parameters, but its predictive performance is still worse.

\begin{table}[]
	\centering
	\renewcommand\arraystretch{1.26}
	\small
	\caption{Ablation studies conducted on training strategies using the KTH, Caltech and KITTI dataset, respectively. The metrics are averaged over the 10 predicted frames.}
	\label{Tab:Ablation}
	\resizebox{\linewidth}{!}
	{
		\begin{tabular}{lccccc}
			\hline
			\multirow{2}{*}{Datasets}  & \multirow{2}{*}{Adv.} & \multirow{2}{*}{Long-term} & \multicolumn{3}{c}{10 $\rightarrow$ 20 }                             \\
			&          &                  & SSIM $\uparrow$          & PSNR $\uparrow$             & LPIPS $\downarrow$           \\ \hline
			\multirow{3}{*}{KTH}      & $\checkmark$ &  $\checkmark$                   & 0.881          & 31.87          & \textbf{7.98}  \\
			&  & $\checkmark$                      & \textbf{0.896} & \textbf{32.43} & 12.19          \\
			&  &                   & 0.892          & 32.33          & 13.26          \\ \hline
			\multirow{3}{*}{Caltech} & $\checkmark$ & $\checkmark$ &                             0.750          & 21.88          & 15.27          \\
			  &  &     $\checkmark$                 & \textbf{0.756} & \textbf{21.95} & \textbf{14.88} \\
			&  &                  & 0.723          & 21.12          & 17.59          \\ \hline
			\multirow{3}{*}{KITTI}    & $\checkmark$ & $\checkmark$                              & 0.538          & 17.57          & 39.31          \\
			&  & $\checkmark$                      & \textbf{0.552} & \textbf{17.99} & \textbf{38.19} \\
			&  &                  & 0.495          & 15.77          & 40.01         \\ \hline
		\end{tabular}
	}
\end{table}

\begin{figure}[]
	\centering{\includegraphics[width=0.49\textwidth]{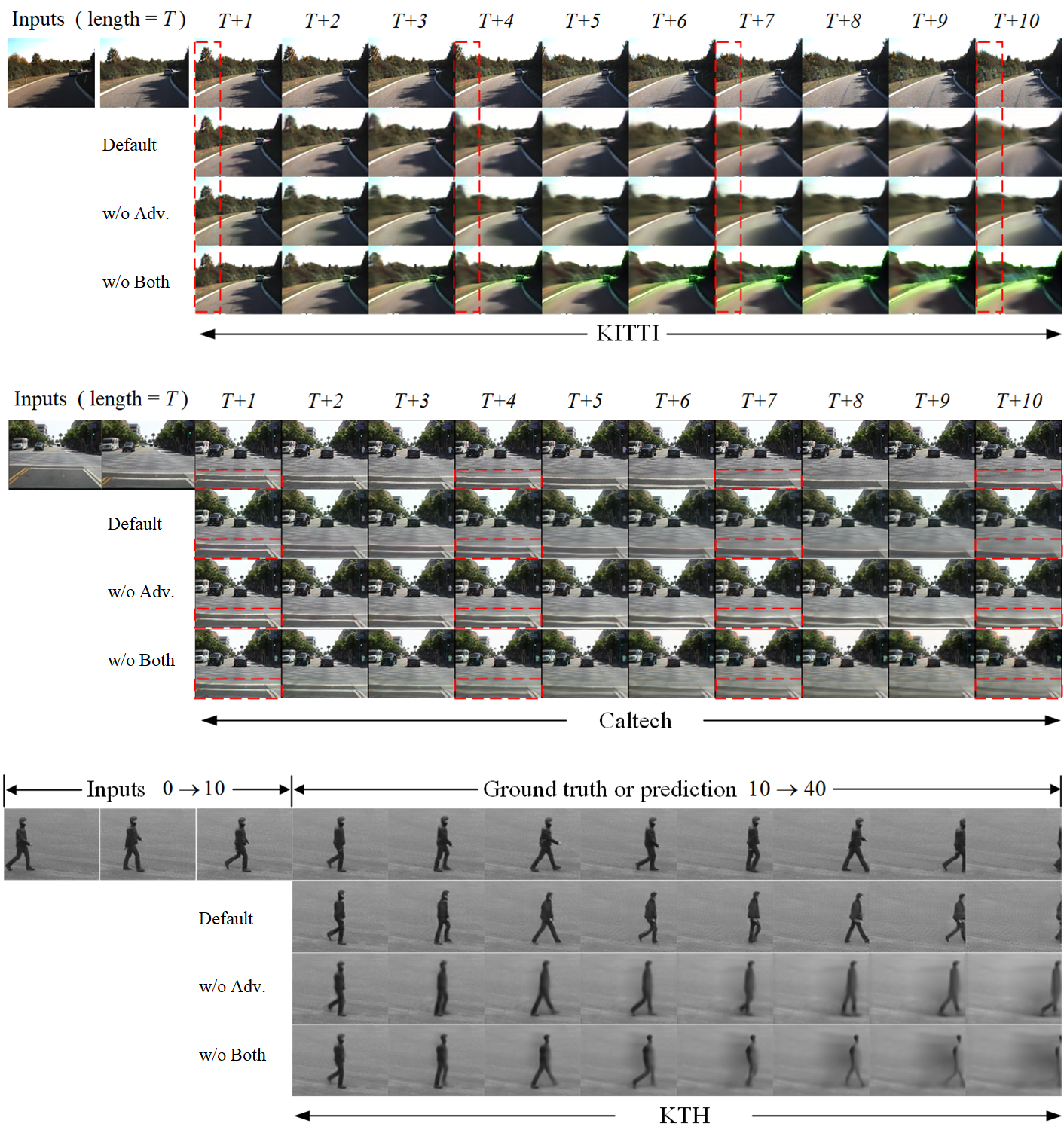}}
	\caption{Visualization examples of ablation studies conducted on training strategies using the KTH, Caltech and KITTI datasets, respectively. We use 10 frames as input to predict the next 10 frames (Caltech and KITTI) or 30 frames (KTH). Zoom in for a better view.}
	\label{fig:Ablation}
\end{figure}

\textbf{Training Strategies} \quad We have observed that training strategies have a substantial impact on the quality of predictions, particularly in terms of visualizing predicted frames. To investigate this influence, we conducted several comparative experiments on the KTH, Caltech, and KITTI datasets. Table \ref{Tab:Ablation} and Figure \ref{fig:Ablation} present the quantitative and qualitative results with different training strategies on the three datasets, respectively. In the table or figure, ``Adv." denotes the improved adversarial training approach, ``Long-term" denotes the strategy explained in Section \uppercase\expandafter{\romannumeral3} B Long-term Prediction, ``w/o Both" denotes that the above two approaches  are not used.

Compared to ``w/o Both", using the ``Long-term" strategy yields better results in both quantitative and qualitative evaluation. This effect is particularly noticeable when dealing with more complex scenes and inter-frame changes, such as in the Caltech and KITTI datasets. In such cases, the difficulty of prediction increases, resulting in faster accumulation of prediction errors. Consequently, the conventional training method (w/o Both) experiences a significant drop in prediction accuracy. On the other hand, employing the "Long-term" strategy enables better suppression of the impact of prediction errors, leading to improved performance under these challenging conditions.

However, although using the ``Long-term" strategy can slow down the degradation of prediction quality, it still suffers from the problem caused by training with deterministic loss—vagueness. Both Euclidean distance and mean square error tend to minimize the loss by averaging, which can easily blur the generated images. Therefore, we introduce the aforementioned adversarial training style to sharpen the generated images. Although its accuracy is somewhat worse, it seems to generate more believable predictions from the perspective of human perception evaluation.

\section{Conclusion}
In this paper, we combined the predictive coding theory and coarse-to-fine approach to design a multi-scale predictive model for the task of video prediction. The total framework of our model is inspired by the computational style of predictive coding. However, traditional predictive coding models only predict what is happening rather than the future, so we proposed to employ a multi-scale approach to improve the architecture. In terms of network modules, we proposed a novel EDLSTM module, which outperforms traditional Encoder-LSTM-Decoder framework in prediction performance while using fewer parameters.
In addition, we made several improvements to the training strategies to alleviate the difficulty in adversarial training and achieve better long-tern prediction.
Importantly, we did get great performance on the task of video prediction with our methods. Nevertheless, the performance of our model and methods show powerless in complex scenarios. There is still huge room for improvement, we will futher explore better methods to make predictions with higher accuracy and clarity in more complex scenarios.

\section{Acknowledgment}
This work was supported in part by the Key-Area Research and Development Program of Guangdong Province under Grant: 2019B090912001, and in part by the Germany/Hong Kong Joint Research Scheme sponsored by the Research Grants Council of Hong Kong and the German Academic Exchange Service of Germany (Ref. No. G-PolyU505/22), PolyU Start-up Grant: ZVUY-P0035417, CD5E-P0043422 and WZ09-P0043123.

\bibliographystyle{IEEEtran}
\bibliography{references}

\end{document}